\documentclass[sigconf]{acmart}
\renewcommand\footnotetextcopyrightpermission[1]{} 
\usepackage{multirow}
\usepackage{enumitem}
\usepackage{graphicx}
\begin{document}

\title[ACPO: Anchor-Constrained Perceptual Optimization for Diffusion Models]{ACPO: Anchor-Constrained Perceptual Optimization for Diffusion Models with No-Reference Quality Guidance}

\author{Yang Yang}
\affiliation{%
  \institution{School of Electronic and Information Engineering, Anhui University}
  \city{Hefei}
  \state{Anhui}
  \country{China}
}
\affiliation{%
  \institution{Institute of Artificial Intelligence, Hefei Comprehensive National Science Center}
  \city{Hefei}
  \state{Anhui}
  \country{China}
}
\email{sky_yang@ahu.edu.cn}

\author{Feifan Meng}
\affiliation{%
  \institution{School of Electronic and Information Engineering, Anhui University}
  \city{Hefei}
  \state{Anhui}
  \country{China}
}
\email{p24301232@stu.ahu.edu.cn}

\author{Han Fang}
\authornote{Corresponding authors.(Email: fanghan@ustc.edu.cn,  zhangwm@ustc.edu.cn)}
\affiliation{%
  \institution{School of Cyber Science and Technology, University of Science and Technology of China}
  \city{Hefei}
  \state{Anhui}
  \country{China}
}
\email{fanghan@ustc.edu.cn}

\author{Weiming Zhang}
\authornotemark[1]
\affiliation{%
  \institution{School of Cyber Science and Technology, University of Science and Technology of China}
  \city{Hefei}
  \state{Anhui}
  \country{China}
}
\email{zhangwm@ustc.edu.cn}
\renewcommand{\shortauthors}{Yang et al.}

\begin{abstract}
  Diffusion models have achieved remarkable success in image generation, yet their training is predominantly driven by full-reference objectives that enforce pixel-wise similarity to ground-truth images.Such supervision, while effective for fidelity, may insufficient in terms of subjective visual perception quality and text-image semantic consistency. In this work, we investigate the problem of incorporating no-reference perceptual quality into diffusion training. A key challenge is that directly optimizing perceptual signals, such as those provided by no-reference image quality assessment (NR-IQA) models, introduces a mismatch with the original diffusion objective, leading to training instability and distributional drift during fine-tuning. To address this issue, we propose an anchor-constrained optimization framework that enables stable perceptual adaptation. Specifically, we leverage a learned NR-IQA model as a perceptual guidance signal, while introducing an anchor-based regularization that enforces consistency with the base diffusion model in terms of noise prediction. This design effectively balances perceptual quality improvement and generative fidelity, allowing controlled adaptation toward perceptually favorable outputs without compromising the original generative behavior. Extensive experiments demonstrate that our method consistently enhances perceptual quality while preserving generation diversity and training stability, highlighting the effectiveness of anchor-constrained perceptual optimization for diffusion models.
\end{abstract}




\maketitle

\makeatletter
\let\ps@ACM\relax      
\def\ps@fancy{}
\def\ps@headings{%
  \let\@oddfoot\@empty
  \let\@evenfoot\@empty
  \def\@oddhead{\normalfont\itshape ACPO: Anchor-Constrained Perceptual Optimization\hfill Yang et al.}
  \def\@evenhead{\normalfont\itshape Yang et al.\hfill ACPO: Anchor-Constrained Perceptual Optimization}
}
\pagestyle{headings}
\makeatother

\section{Introduction}

    \begin{figure}[h] 
		\centering 
		\includegraphics[width=0.95\linewidth]{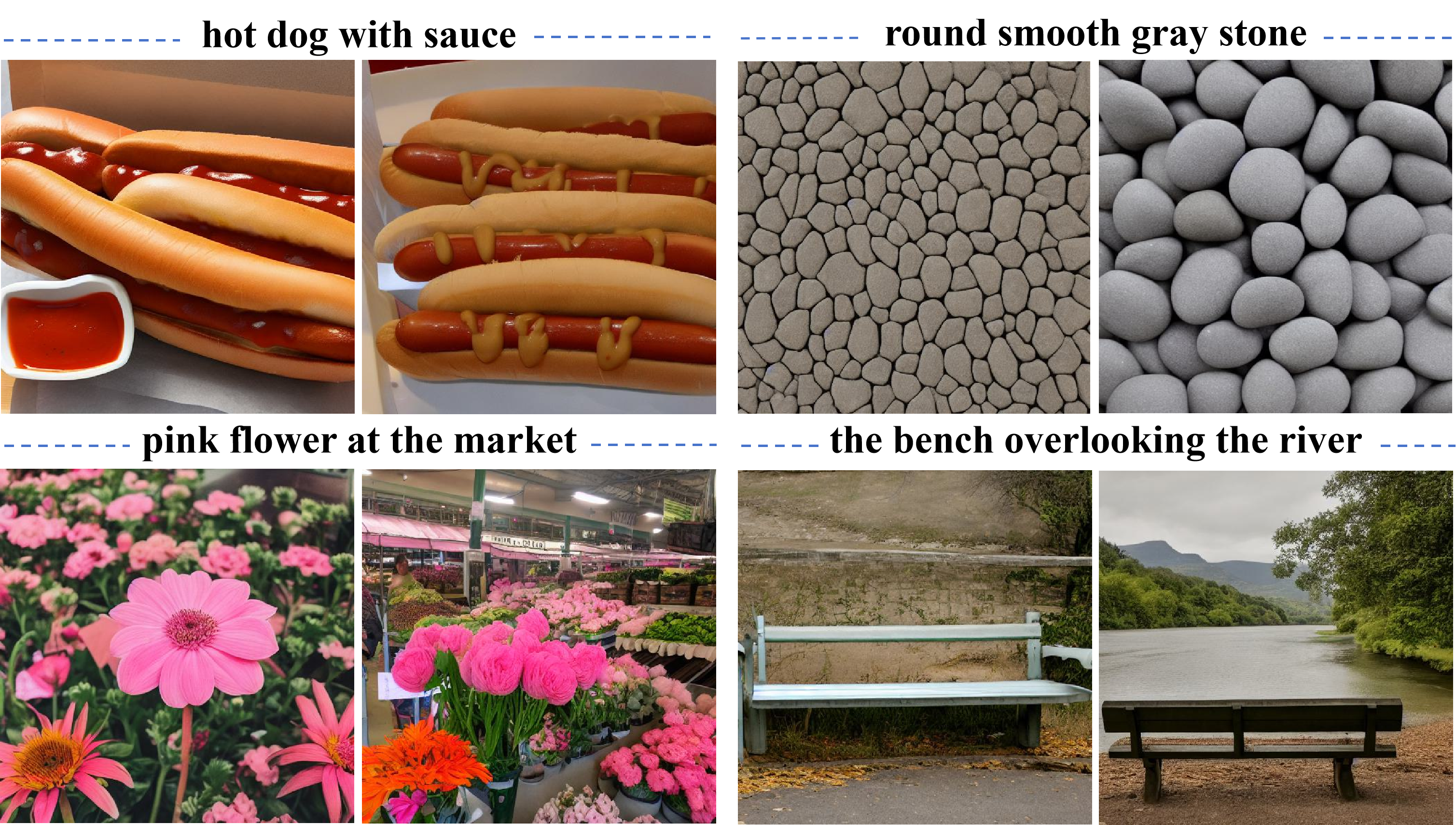} 
		\caption{Teaser. Comparison between the baseline model (left) and our method (right). After fine-tuning with our framework, the generated images show improved semantic consistency and perceptual quality.}
		\Description{Four pairs of side-by-side image generation results are shown, with baseline outputs on the left and our method's outputs on the right. Each pair corresponds to a text prompt, our method's outputs better align with the semantic meaning of each prompt: clearer subject focus, more accurate scene composition, and higher perceptual fidelity compared to the baseline results}
        \label{fig:comparison}
    \end{figure}
    
Generative models have become a fundamental component of modern computer vision, enabling machines to synthesize realistic images and to model complex data distributions. Over the past decade, significant progress has been made in generative modeling, leading to remarkable advances in image synthesis, editing, and translation. Among various generative paradigms, diffusion models have recently emerged as a powerful and versatile framework, demonstrating state-of-the-art performance in high-fidelity image generation across a wide range of tasks\cite{Ho2020DDPM,ho2022cascaded}. By modeling the data distribution through a forward noise injection process and a learned reverse denoising process, diffusion models are capable of generating visually plausible images with strong diversity and stability\cite{Song2021ScoreBased}. Compared to adversarial learning frameworks\cite{Krichen2023gans}, diffusion-based methods exhibit more stable training dynamics and have achieved state-of-the-art performance across a wide range of image generation tasks, including unconditional generation, conditional synthesis, and image-to-image translation.

Despite these remarkable advances, the training of most diffusion models is predominantly driven by full-reference objectives, typically formulated as a pixel-wise mean squared error (MSE) between the predicted noise and the ground-truth noise. While such objectives are mathematically convenient and highly effective for ensuring basic generation fidelity and distribution matching, their alignment with subjective human visual perception remains suboptimal \cite{wang2004SSIM,zhang2018LPIPS}. As a result, models trained wit
h only pixel-level losses, while capable of producing visually plausible results that align well with reference data, may still leave room for further improvement in subjective visual quality and text-image semantic alignment.

This observation highlights a promising direction for refining current diffusion training pipelines. Pixel-level objectives excel at encouraging accurate structural and distributional alignment with ground-truth data, yet achieving close pixel-level similarity does not always fully translate to optimal perceptual appeal and semantic coherence as perceived by humans. Therefore, incorporating no-reference image quality assessment (NR-IQA) signals into fine-tuning is a natural progression. However, while introducing NR-IQA metrics offers clear potential for further improvement, directly embedding such guidance throughout the full training process presents notable practical hurdles. Directly optimizing an NR-IQA-guided signal fundamentally mismatches the original noise-prediction objective. In practice, this misalignment causes severe training instability and distributional drift, often compromising the model's original generative capabilities in pursuit of higher IQA scores. This raises a crucial question: how can we effectively integrate no-reference perceptual quality into diffusion fine-tuning without disrupting the stability and generative behavior of the base model?

To answer this question, we propose Anchor-Constrained Perceptual Optimization (ACPO), a novel fine-tuning framework that safely and effectively integrates NR-IQA feedback into diffusion models. Instead of naively applying perceptual guidance across the entire reverse process, ACPO establishes an iterative optimization loop specifically targeting the late stages of denoising, where fine-grained structural details and textures typically solidify. Crucially, to prevent distributional drift, this perceptual guidance is coupled with an anchor-based regularization mechanism. By penalizing the noise prediction deviation between lightweight trainable modules and the frozen pre-trained backbone, ACPO softly constrains the model to its original denoising dynamics. This dual-objective design ensures that the model can progressively enhance subjective visual quality and semantic alignment without compromising its inherent stability and generative diversity. Consequently, explicitly optimizing for the NR-IQA feedback allows the generated model to synthesize images with richer details that better match human perceptual quality, as demonstrated in Figure~\ref{fig:comparison} (top row). Notably, when the guided NR-IQA metric encapsulates semantic attributes, ACPO further empowers the model to achieve a substantial improvement in text-image alignment (Figure~\ref{fig:comparison}, bottom row).

The main contributions of this work are summarized as follows: 
\begin{enumerate}[label=\arabic*), leftmargin=*, itemsep=0.2em, topsep=0.2em, partopsep=0pt, parsep=0pt]	
\item We propose Anchor-Constrained Perceptual Optimization (ACPO), a general quality-metric-driven training framework that introduces no-reference perceptual quality signals into diffusion model fine-tuning without requiring reference images or additional annotations.

\item An effective quality-guided loss formulation, coupled with an anchor-based regularization mechanism, is designed to enable perceptual signals to safely guide diffusion training while strictly preserving training stability and generation diversity.

\item Extensive experiments across multiple datasets, image resolutions, and diffusion configurations demonstrate that the proposed approach achieves consistent improvements in both perceptual quality metrics and text-image semantic consistency.
\end{enumerate}


\section{Related Works}
\subsection{Image Quality Assessments}
Image Quality Assessment (IQA) methods are typically categorized into three types based on the availability of reference information: fully referenced (FR-IQA), partially referenced (RR-IQA), and non-referenced (NR-IQA). FR-IQA requires a "distortion-free reference image" corresponding to the image under assessment, RR-IQA uses only partial reference information, while NR-IQA predicts perceptual quality directly from the input image without relying on reference images.

\subsubsection{Full-Reference Image Quality Assessments}
FR-IQA evaluates the perceptual quality of a distorted image by explicitly comparing it with a pristine reference. Early pixel-wise metrics like MSE and PSNR correlate poorly with human perception, as they ignore structural and contextual distortions. To address this, SSIM \cite{wang2004SSIM} introduced structural similarity modeling, paving the way for methods exploring richer perceptual features. For instance, FSIM \cite{zhang2011FSIM} and GMSD \cite{xue2014GMSD} incorporate gradient information and phase congruency to capture local structural variations, while VSI \cite{zhang2014VSI} integrates visual saliency models to align with human perceptual focus.With the advent of deep learning, the field has increasingly adopted learned representations. Recent models such as DISTS \cite{ding2020DISTS} and LPIPS-ViT \cite{zhang2018LPIPS} leverage pretrained vision transformers (ViTs) to capture perceptual fidelity at higher semantic levels. Additionally, Transformer-based NR-IQA methods like ViT-IQA \cite{You2021TransformerIQ} demonstrate how transformers can jointly capture low-level visual fidelity and high-level semantic structures. Collectively, these developments mark a clear evolution from simple pixel-level metrics toward deep models that better evaluate complex generative distortions.

\subsubsection{No-Reference Image Quality Assessments}
In practice, it is difficult to obtain the reference image in many applications, which makes the No-Reference Image Quality Assessment (NR-IQA) especially important in the context of generative image modeling tasks. Early NR-IQA methods relied on natural scene statistics (NSS) to evaluate image naturalness and local distortions, as seen in BRISQUE \cite{Mittal2012BRISQUE} and the fully blind NIQE \cite{Mittal2013NIQE}. However, these statistical approaches struggle with the complex artifacts introduced by modern generative models.
The advent of deep learning significantly advanced the field. Methods like Two-Stream CNN \cite{Yan2019TwoStream} leveraged gradient maps for structural details, while HyperIQA \cite{Su2020HyperIQA} utilized hypernetworks to dynamically adapt to diverse distortions. To specifically address complex generative artifacts, researchers also explored adversarial inference paradigms \cite{Ma2020ActiveInferenceGAN}, using GANs to generate internal pseudo-references for robust quality prediction.
More recently, the focus has shifted toward Transformer architectures, semantic alignment, and human perception \cite{li2025scagiqa}. For instance, MANIQA \cite{Yang2022MANIQA} employs a multi-scale Transformer to assess complex generative distortions like structural and color misalignment. For text-to-image tasks, evaluating text-image consistency has become vital; IPCE \cite{Peng2024IPCE} specifically estimates image-prompt correspondence, while SF-IQA \cite{Yu2024SFIQA} leverages Swin Transformers and CLIP for robust semantic quality assessment. Furthermore, models like PickScore \cite{Kirstain2023PickAPic} directly align quality predictions with user expectations by training on large-scale human preference data. As highlighted in recent literature \cite{Tian2025SurveyT2IQA}, these advancements in visual coherence and prompt alignment are now central to evaluating modern generative models.

\subsection{Generative Models}

\begin{figure*}[t]
    \centering
    \includegraphics[width=\textwidth]{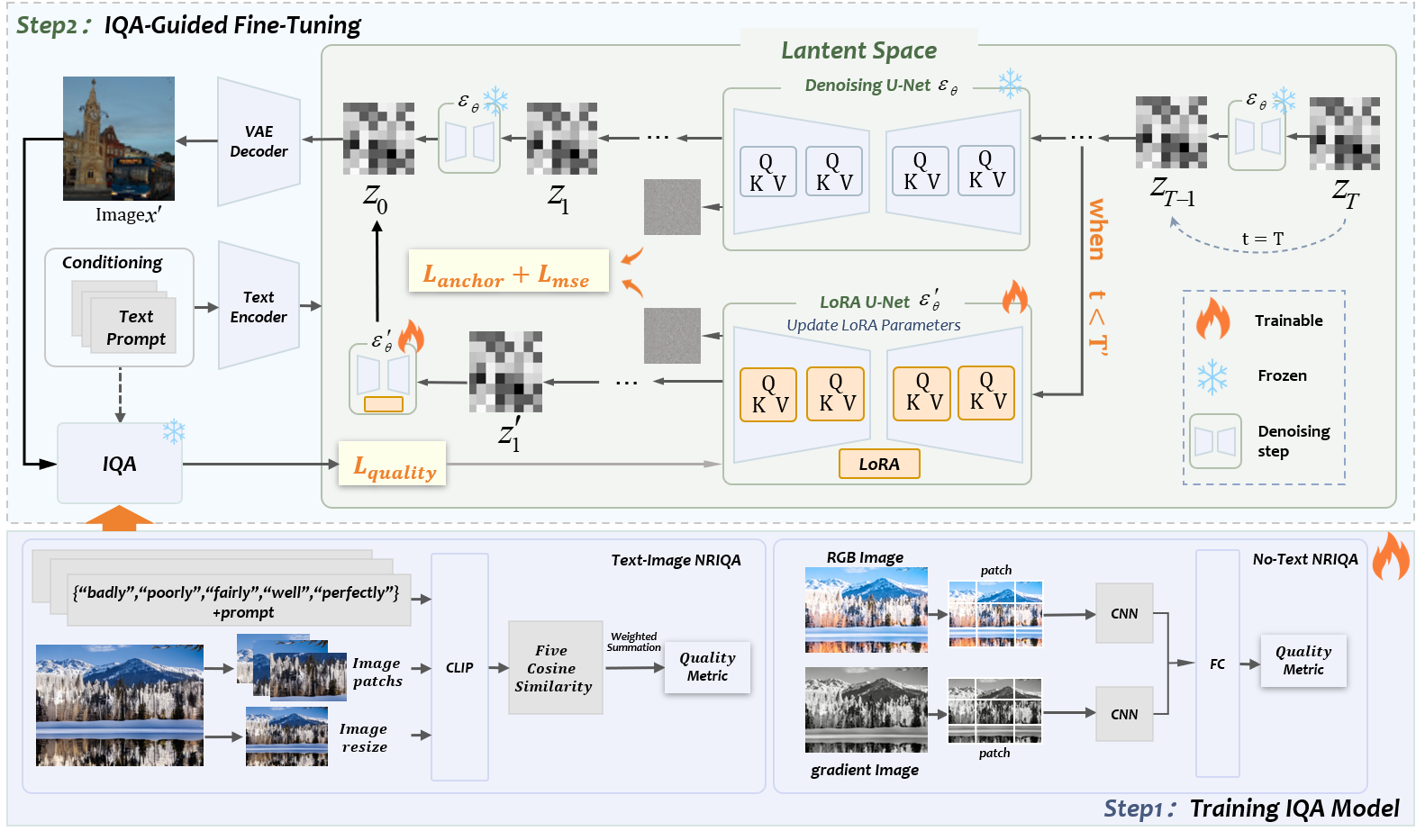}
    \caption{Overview of the Anchor-Constrained Perceptual Optimization (ACPO) framework. 
    The framework integrates diffusion-based image generation with LoRA adaptation, perceptual quality assessment, and quality-aware regularization into a unified iterative optimization pipeline. Notably, perceptual feedback is dynamically provided based on the generation task: a standard NR-IQA model~\cite{Yan2019TwoStream} is utilized for unconditional settings, while a CLIP-based IQA model~\cite{Peng2024IPCE} evaluates text-conditional generation. The resulting task-specific quality feedback iteratively guides the parameter updates.}
    \Description{Four comparative images of the generated church.}
    \label{fig:framework}
\end{figure*}

The development of deep generative models has progressed significantly over the past decade. Early paradigms, including Variational Autoencoders (VAEs) \cite{Kingma2014VAE} and autoregressive models \cite{Oord2016PixelRNN}, pioneered probabilistic likelihood modeling but often struggled with overly smooth outputs or high computational costs. Generative Adversarial Networks (GANs) \cite{Radford2016DCGAN} subsequently achieved high visual realism, though they remained notoriously difficult to optimize. A major paradigm shift occurred with diffusion models, which formulate generation as a learned denoising process. Denoising Diffusion Probabilistic Models (DDPM) \cite{Ho2020DDPM} demonstrated superior sample quality without adversarial instability. This foundation was rapidly advanced by DDIM \cite{Song2021DDIM} for accelerated sampling, and score-based formulations with stochastic differential equations (SDEs) \cite{Song2021SDE}, which provided a unified continuous-time theoretical framework.To enhance scalability and versatility, subsequent research focused on efficiency and high-resolution synthesis. Cascaded architectures \cite{Ramesh2022HierarchicalCLIPLatents} and Latent Diffusion Models (LDMs) \cite{Rombach2022LDM} drastically reduced computational costs by performing diffusion in a compressed latent space. This catalyzed the deployment of large-scale text-to-image systems like Imagen \cite{Saharia2022Imagen}, which exhibited strong semantic alignment. Concurrently, frameworks such as ControlNet \cite{Zhang2023ControlNet} enabled fine-grained, controllable generation guided by external structural constraints.Recently, diffusion-based generation has evolved toward Transformer-based backbones (DiTs) and unified creation frameworks \cite{yang2024hi3d}. PixArt-$\Sigma$ \cite{Chen2024PixArtSigma} and Hunyuan-DiT \cite{Li2024HunyuanDiT} leverage DiT architectures to achieve highly efficient, 4K-resolution synthesis, emphasizing strong prompt adherence and fine-grained multilingual semantic understanding. Beyond pure synthesis, models like ACE \cite{Han2024ACE} extend these capabilities into instruction-following frameworks, supporting diverse generation and editing scenarios within a single unified architecture.

Collectively, these developments underscore the rapid evolution of diffusion models in resolution scalability, semantic alignment, controllability, and task generalization—while also revealing that most training pipelines remain dominated by pixel-aligned objectives, motivating perceptually aware optimization strategies.

\section{Method}

\subsection{Motivation and Overview}
As established, naively injecting no-reference image quality assessment (NR-IQA) signals directly into the diffusion process severely disrupts the pre-trained denoising dynamics. Rather than the model genuinely improving visual quality, it often exploits adversarial shortcuts to artificially inflate IQA scores, leading to mode collapse and severe distribution drift. Therefore, the core technical challenge lies in designing a stable iterative optimization scheme that leverages perceptual feedback while strictly constraining these adversarial deviations.

To address this limitation, we propose the Anchor-Constrained Perceptual Optimization (ACPO) framework to safely integrate NR-IQA feedback into diffusion training. As illustrated in Figure~\ref{fig:framework}, the overall pipeline of our framework consists of two main stages. The first stage focuses on training a differentiable NR-IQA evaluator tailored to the specific generative task. For unconditional diffusion models, the evaluator is trained exclusively to capture human visual preferences and perceptual quality. For text-conditioned models, the training objective is expanded to assess both visual fidelity and text-image semantic alignment.

In the second stage, we leverage this trained evaluator to guide the fine-tuning of the diffusion model. Specifically, intermediate samples synthesized during the late denoising stages are evaluated by the NR-IQA model to provide a perceptual guidance signal. Crucially, to maximize this perceptual score without causing distributional drift, the guidance is jointly optimized with an anchor-based regularization provided by the frozen pre-trained backbone. Through a stable iterative optimization scheme, ACPO updates only the lightweight modules (e.g., LoRA) while keeping the base diffusion backbone and the evaluator strictly frozen, thereby progressively enhancing perceptual quality while strictly preserving the original generative prior.

\subsection{Differentiable NR-IQA Evaluator Preparation}
To provide accurate perceptual guidance, the first stage of ACPO establishes a differentiable NR-IQA evaluator. Depending on the generative task, we tailor the evaluator to ensure the scoring mechanism perfectly aligns with the target modality and provides a reliable optimization signal. 

For text-conditioned generation, we explicitly model both semantic consistency and structural coherence. Specifically, we follow the evaluation paradigm established in IPCE \cite{Peng2024IPCE} to train the NIQA model, which explicitly quantifies text-image alignment and internal structural fidelity. Given a generated image $\hat{x}$ and text condition $c$, we extract their CLIP representations $f_{\text{img}}(\hat{x})$ and $f_{\text{text}}(c)$. The semantic alignment score $S_{\text{sem}}$ evaluates their cosine similarity in the joint embedding space:
\begin{equation}
S_{\text{sem}}(\hat{x}, c) = \frac{\langle f_{\text{img}}(\hat{x}), f_{\text{text}}(c) \rangle}{\|f_{\text{img}}(\hat{x})\|_2 \, \|f_{\text{text}}(c)\|_2}.
\end{equation}
To capture internal structural coherence, we compute the agreement between local patch tokens $p_k^{(l)}$ and the global token $g^{(l)}$ across selected CLIP encoder layers $\mathcal{L}$, weighted by $\alpha_l$:
\begin{equation}
S_{\text{str}}(\hat{x}) = \sum_{l \in \mathcal{L}} \alpha_l \left( \frac{1}{K_l} \sum_{k=1}^{K_l} \frac{\langle p_k^{(l)}, g^{(l)} \rangle}{\|p_k^{(l)}\|_2 \, \|g^{(l)}\|_2} \right).
\end{equation}
To ensure numerical stability and bounded optimization signals, the final perceptual score is formulated as a sigmoid-normalized linear combination of these metrics:
\begin{equation}
\tilde{Q}(\hat{x}, c) = \text{sigmoid}\left(\beta_1 S_{\text{sem}}(\hat{x}, c) + \beta_2 S_{\text{str}}(\hat{x}) + \beta_3\right),
\end{equation}
where $\beta_i$ are balancing coefficients.

For unconditional generation where textual guidance is absent, semantic alignment is inapplicable. Instead, we train a two-stream NR-IQA architecture \cite{Yan2019TwoStream} to capture both appearance degradation and structural distortions. This network utilizes parallel encoders $\Phi_{\text{rgb}}$ and $\Phi_{\text{grad}}$ to extract features from the raw image $\hat{x}$ and its high-frequency gradient map $\nabla \hat{x}$, respectively, which effectively highlights subtle edge artifacts. These representations are concatenated and mapped through a fusion network $\mathcal{F}_{\text{fuse}}$ (e.g., a multi-layer perceptron) to yield a bounded perceptual score:
\begin{equation}
\tilde{Q}(\hat{x}) = \text{sigmoid}\left( \mathcal{F}_{\text{fuse}}\left( [\Phi_{\text{rgb}}(\hat{x}), \Phi_{\text{grad}}(\nabla \hat{x})] \right) \right),
\end{equation}
where $[\cdot, \cdot]$ denotes concatenation.

\subsection{Anchor-Constrained IQA-Guided Fine-Tuning}
With the differentiable NR-IQA evaluator established in the first stage, the second stage integrates this perceptual feedback into the diffusion fine-tuning process. Rather than treating quality assessment as a passive, post-training metric, we elevate it to an active driving force and organize this phase as a tightly coupled, iterative optimization procedure. Specifically, to achieve efficient adaptation without catastrophic forgetting, we insert lightweight LoRA modules ($\theta$) into the pre-trained diffusion backbone, which remains strictly frozen alongside the IQA evaluator.

\subsubsection{Targeted Generation and Perceptual Guidance}
This iterative optimization explicitly targets the late denoising stages. In standard diffusion models, the reverse generation process is highly stage-dependent: early timesteps (associated with high noise levels) primarily establish coarse semantic layouts and global compositions, while late timesteps (approaching $t=0$) are responsible for solidifying fine-grained structural details, lighting, and textures—precisely the visual attributes that dictate human perceptual quality.

Therefore, rather than applying guidance uniformly across all $t \in [1, T]$, we restrict perceptual updates to a late timestep range $\mathcal{T}_{\text{late}}$. This avoids disrupting the global layouts established in early stages, allowing the model to safely refine high-frequency details without corrupting the composition. Given a textual condition $c$ and a latent variable $z$, the LoRA-equipped generator $G_{\theta}$ progressively denoises the representation within these targeted timesteps. The spatial decoder $D(\cdot)$ then projects this representation back to pixel space to synthesize an intermediate image:
\begin{equation}
\hat{x} = D\left(G_{\theta}(c, z)\right).
\end{equation}

To actively drive optimization, this generated sample $\hat{x}$ is immediately routed into the frozen NR-IQA model from Stage 1. This yields a stable, differentiable perceptual score $\tilde{Q}(\hat{x}, c) \in (0, 1)$. Since a higher score indicates superior visual coherence and text-image semantic alignment, we formulate a perceptual quality-guided loss. This objective explicitly penalizes the discrepancy between the current generated quality and the ideal upper bound. During backpropagation, gradients derived from this loss flow seamlessly through the frozen evaluator and spatial decoder to update the trainable parameters $\theta$, compelling the network to iteratively refine its outputs:
\begin{equation}
\mathcal{L}_{\text{quality}}(\theta) = \mathbb{E}_{c \sim \mathcal{C},\, z \sim p(z)} \left[ 1 - \tilde{Q}\left(\hat{x}, c\right) \right].
\label{eq:qual_loss}
\end{equation}

\subsubsection{Anchor-Based Regularization}
While minimizing $\mathcal{L}_{\text{quality}}$ provides powerful guidance for improving perceptual quality, relying solely on this objective introduces a critical vulnerability. Trainable parameters tend to exploit adversarial shortcuts within the differentiable evaluator to artificially inflate the score. This phenomenon inevitably disrupts the pre-trained denoising dynamics, leading to severe mode collapse and significant distributional drift.

To safely harness perceptual guidance without compromising the model's inherent generative prior, we introduce an anchor-based regularization mechanism. This mechanism mathematically bounds the optimization trajectory by treating the frozen base model's predictions as a stabilizing anchor. Specifically, over the targeted late denoising timestep range $\mathcal{T}_{\text{late}}$, we softly constrain the noise prediction of the trainable network $\epsilon_\theta$ to closely track the concurrent prediction of the frozen base model $\epsilon_{\text{base}}$:
\begin{equation}
\mathcal{L}_{\text{anchor}}(\theta) = \mathbb{E}_{x_t, c, t \in \mathcal{T}_{\text{late}}} \left[ \|\epsilon_\theta(x_t, t, c) - \epsilon_{\text{base}}(x_t, t, c)\|_2^2 \right],
\label{eq:anchor_loss}
\end{equation}
where $x_t$ represents the noisy latent at timestep $t$. This dual-objective formulation grants the LoRA modules flexibility to enhance perceptual details via $\mathcal{L}_{\text{quality}}$, while strictly preventing excessive deviation from the original, well-behaved data distribution.

\subsubsection{Stable Iterative Optimization Procedure}
To guarantee fundamental generative fidelity and retain basic text-to-image mapping capabilities, 
we maintain $\mathcal{L}_{\text{mse}}$, the original denoising MSE loss of the diffusion model, throughout the optimization process:
\begin{equation}
\mathcal{L}_{\text{mse}}(\theta) = \mathbb{E}_{x_0, \epsilon, t, c} \left[ \|\epsilon - \epsilon_{\theta}(x_t, t, c)\|_2^2 \right].
\end{equation}

The comprehensive training objective for this iterative optimization procedure is then formulated as a principled combination of the standard diffusion loss, the structural anchor constraint, and the explicit perceptual guidance:
\begin{equation}
\mathcal{L}_{\text{total}}(\theta) = \mathcal{L}_{\text{mse}}(\theta) + \lambda_1 \mathcal{L}_{\text{anchor}}(\theta) + \lambda_2 \mathcal{L}_{\text{quality}}(\theta),
\label{eq:total_loss}
\end{equation}
where $\lambda_1$ and $\lambda_2$ are hyperparameters governing the regularization strength and the magnitude of perceptual enhancement, respectively. 

In practice, the ACPO framework operates through a tightly coupled evaluation-update iteration. During each step, the generator produces a targeted late-stage output, the frozen IQA model immediately evaluates it to provide gradient feedback, and the frozen backbone concurrently supplies the reference noise for structural regularization. By strictly updating only the lightweight parameters during this iterative optimization, we safely translate the static evaluator from Stage 1 into an active, robust driving force for perceptual alignment.

\section{Experiments}

\subsection{Experimental Setup}
\begin{table*}[t]
  \caption{Comparison of DDPM with and without Quality-Guided Training Across Datasets at Various Resolutions}
  \label{tab:ddpm_main_results}
  \centering
  \setlength{\tabcolsep}{6pt}
  \begin{tabular}{llccccc}
    \toprule
    \textbf{Dataset} & \textbf{Method} & \textbf{MSE $\downarrow$} & \textbf{TwostreamIQA\cite{Yan2019TwoStream} $\uparrow$} & \textbf{HyperIQA\cite{Su2020HyperIQA} $\uparrow$} & \textbf{MANIQA\cite{Yang2022MANIQA} $\uparrow$} \\
    \midrule
    \multirow{2}{*}{CIFAR-10($32 \times 32$)}
    & DDPM Baseline         & 0.0270 & 78.92 & 0.1924 & 18.28 \\
    & \textbf{Ours (DDPM + Framework)} & \textbf{0.0267} & \textbf{80.94} & \textbf{0.1941} & \textbf{18.41} \\
    \midrule
    \multirow{2}{*}{Anime-Faces($64 \times 64$)}
    & DDPM Baseline         & 0.0186 & 82.53 & 0.2439 & 24.16 \\
    & \textbf{Ours (DDPM + Framework)} & \textbf{0.0187} & \textbf{84.69} & \textbf{0.2464} & \textbf{24.21} \\
    \midrule
    \multirow{2}{*}{LSUN Church($256 \times 256$)}
    & DDPM Baseline         & 0.0179 & 76.63 & 0.5444 & 59.30 \\
    & \textbf{Ours (DDPM + Framework)} & \textbf{0.0177} & \textbf{79.16} & \textbf{0.5401} & \textbf{58.88} \\
    \bottomrule
  \end{tabular}
\end{table*}

\subsubsection{Implementation Details}
We implement our framework in PyTorch and conduct all experiments on a single NVIDIA A800 GPU. For the base generators, we evaluate on DDPM (trained from scratch) and Stable Diffusion v1.4 (SD-v1.4, initialized from official weights). For SD-v1.4, we achieve parameter-efficient fine-tuning by injecting LoRA modules into the U-Net linear layers while keeping the backbone strictly frozen. DDPM is evaluated on CIFAR-10 \cite{Krizhevsky2009LearningML}, Anime-faces \cite{AnimeFacesDataset} and LSUN datasets \cite{Yu2015ConstructionOA}. SD-v1.4 is fine-tuned using prompts sampled from Visual Genome \cite{Krishna2017VisualGenome}, ranging from single category prompts to diverse large-scale sets, to verify generalization. The guiding NR-IQA model is pre-trained via a three-stage strategy on AGIQA-1K \cite{li2023agiqa1k}, AGIQA-3K \cite{li2023agiqa3k}, AIGCIQA2023 \cite{wang2023aigciqa2023}, and AIGCQA-30K-Image \cite{li2024aigcqa30k}. During optimization, we train the LoRA parameters using the Adam optimizer with a learning rate of 2e-5 for 1,000 steps under the joint objective.

\subsubsection{Evaluation Metrics}
To comprehensively and objectively evaluate the quality of generated images, we employ two complementary sets of metrics for training guidance and unbiased testing.

\paragraph{Quality-Guided Metrics:} During training, TwoStream-IQA \cite{Yan2019TwoStream} and IPCE \cite{Peng2024IPCE} serve as the differentiable NR-IQA evaluators within our ACPO framework. These metrics explicitly drive the optimization process, guiding the model to enhance perceptual fidelity and structural integrity during fine-tuning.

\paragraph{Independent Evaluation Metrics:} To ensure fair and unbiased evaluation, we assess the generated images using a separate set of unseen metrics. Specifically, we adopt MANIQA \cite{Yang2022MANIQA} and HyperIQA \cite{Su2020HyperIQA} to evaluate pure perceptual image quality, alongside PickScore \cite{Kirstain2023PickAPic} and CLIPScore \cite{Hessel2021CLIPScoreAR} to measure text-image semantic alignment and aesthetic preference.

\subsubsection{Baselines and Compared Methods}
To validate the ACPO framework, we establish strictly consistent evaluation protocols (identical datasets, sample counts, and metrics) and compare our approach against the following configurations: The original DDPM and Stable Diffusion (SD) models. These rely exclusively on the standard noise prediction objective without any explicit perceptual guidance. To isolate the impact of parameter-efficient tuning, we evaluate a variant where LoRA modules are introduced but optimized solely with the standard diffusion loss. This confirms that performance gains stem from our ACPO mechanism rather than merely increased parameter capacity. Our full framework integrates the proposed dual-objective (perceptual quality guidance and anchor regularization) into the fine-tuning process, explicitly driving the model toward higher perceptual fidelity and semantic consistency.

\subsection{Baseline Experimental Results}
\subsubsection{Result on DDPM}

\begin{figure*}[t]
    \centering
    \includegraphics[width=\textwidth]{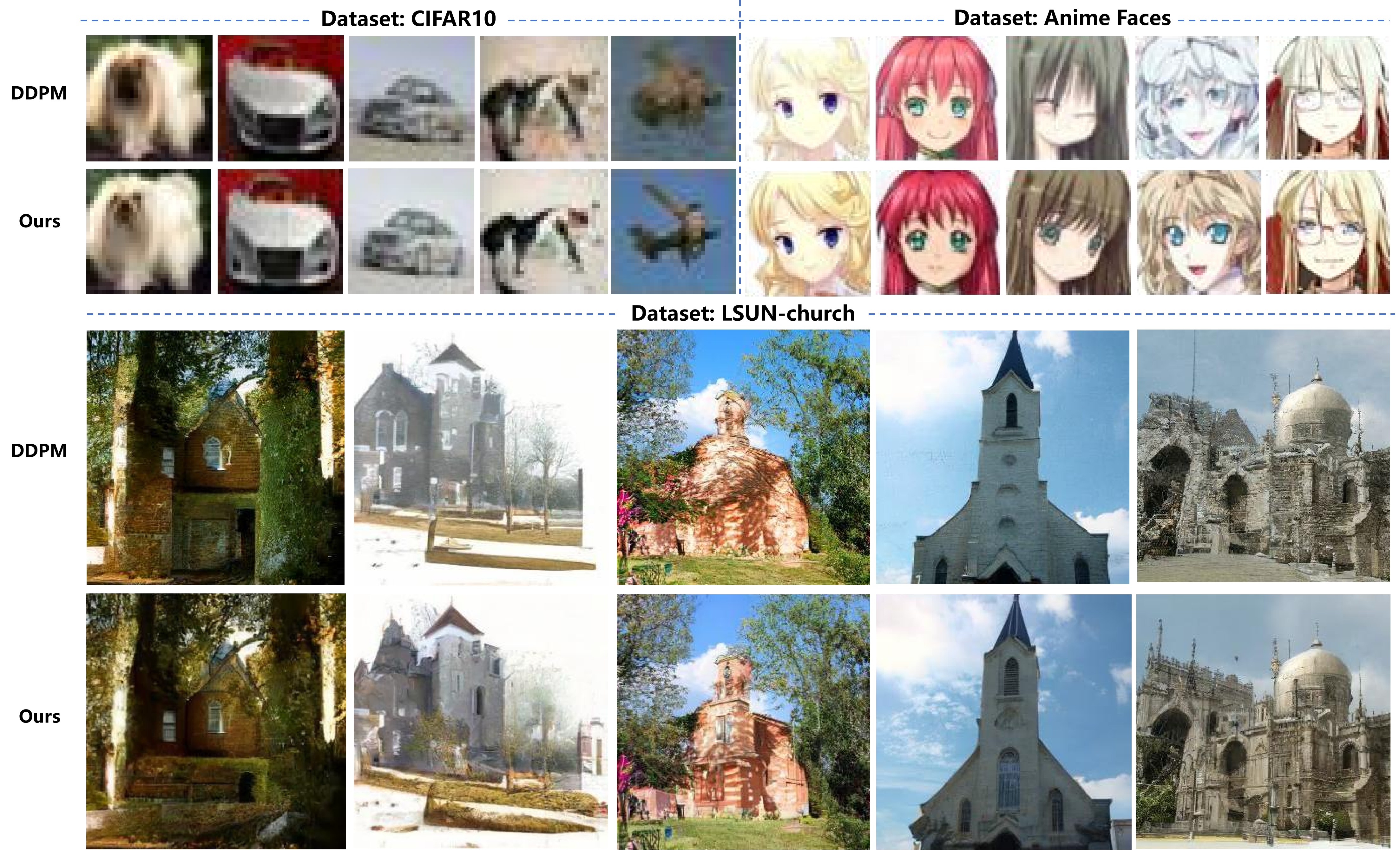}
    \caption{Qualitative comparison between the baseline DDPM and the proposed quality-aware diffusion model across multiple datasets. 
    For each dataset, images in the top row are generated by the baseline DDPM, while the bottom row shows the corresponding results from our method. 
    Each column is generated from the same initial Gaussian noise under identical sampling settings, ensuring a fair comparison.}
    \Description{Three rows of side-by-side image generation results are presented. Each row contains two groups labeled "Baseline" and "Ours". The first row shows real-world objects (dogs, cars, planes), the second row shows anime-style character portraits, and the third row shows architectural scenes (houses, churches, historic buildings). For each sample, the "Ours" results exhibit clearer structures, sharper details, and more coherent textures compared to the blurry or distorted "Baseline" results.}
    \label{fig:visual_results1}
\end{figure*}

Table~\ref{tab:ddpm_main_results} and Figure~\ref{fig:visual_results1} report the quantitative and qualitative comparisons between the baseline DDPM and our ACPO framework across CIFAR-10 ($32 \times 32$), Anime-Faces ($64 \times 64$), and LSUN Church ($256 \times 256$).

\begin{table}[!htbp]
  \caption{Performance comparison between baseline and fine-tuned models on various evaluation metrics.}
  \label{tab:SD_main_results}
  \centering
  \setlength{\tabcolsep}{1pt} 
  \begin{tabular}{lcc|cc|cc}
    \toprule
    \textbf{Metric} & \textbf{Baseline} & \textbf{Fine-tuned} & \textbf{Std.} & \textbf{$t$-statistic} & \textbf{Win Rate} \\
    \midrule
    IPCE\cite{Peng2024IPCE}$\uparrow$       & 2.8831   & 2.9166     & 0.2176 & 3.4424      & 82.4\%   \\
    Pickscore\cite{Kirstain2023PickAPic}$\uparrow$  & 20.5847  & 21.1083    & 0.9112 & 12.8494     & 72.6\%   \\
    CLIPscore\cite{Hessel2021CLIPScoreAR}$\uparrow$  & 0.2840   & 0.2983     & 0.0287 & 11.1089     & 67.2\%   \\
    Twostream\cite{Yan2019TwoStream}$\uparrow$  & 74.9430  & 76.1658    & 11.0330& 2.4783      & 57.6\%   \\   
    \bottomrule
  \end{tabular}
\end{table}

\paragraph{Objective Results:}
As shown in Table~\ref{tab:ddpm_main_results}, ACPO consistently improves perceptual metrics across all resolutions. Notably, TwoStream-IQA scores increase significantly, rising from 78.92 to 80.94 on CIFAR-10 and from 82.53 to 84.69 on Anime-Faces. Similar steady gains are observed in HyperIQA and MANIQA. Crucially, the MSE values remain comparable or slightly lower than the baseline. This confirms that our anchor-constrained optimization effectively enhances perceptual fidelity and maintains stable generative dynamics, firmly preventing distributional drift.

\paragraph{Subjective Results:}
Figure~\ref{fig:visual_results1} corroborates these numerical gains, proving that the ACPO framework successfully steers the diffusion model toward perceptually superior solutions. Evaluated under identical initial Gaussian noise, the baseline outputs frequently suffer from blurry contours, fragmented facial features, and incomplete architectural geometries. In stark contrast, ACPO generates images with significantly sharper boundaries, geometrically coherent structures, and finer texture details. These tangible visual enhancements directly align with our quantitative improvements, demonstrating that integrating NR-IQA feedback achieves optimal perceptual quality without compromising original generative diversity.

\subsubsection{Result on Stable Diffusion}

\begin{figure*}[t]
    \centering
    \includegraphics[width=\textwidth]{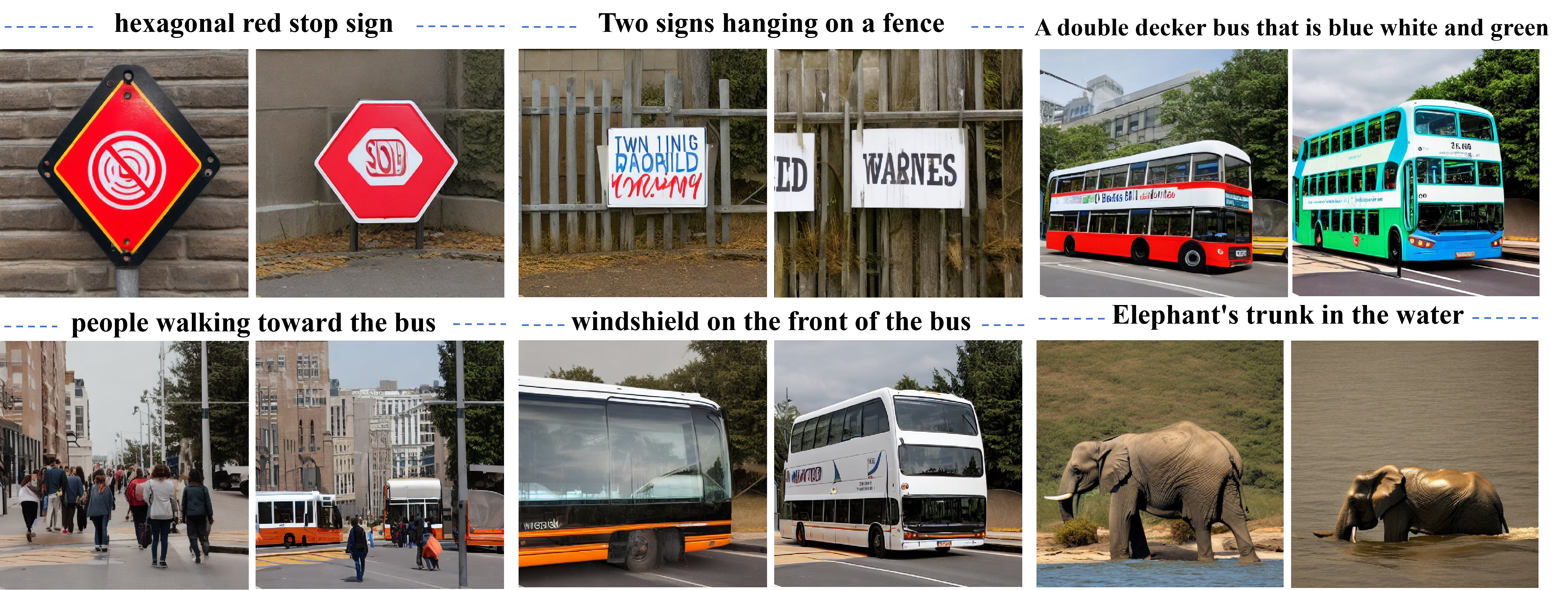}
    \caption{The visual results of text-to-image generation between the baseline Stable Diffusion and our improved model. For each prompt (indicated by the blue dashed labels), the left image shows the result from the baseline model, while the right image displays the result from our proposed method. All image pairs are generated using the same random seed and sampling parameters to ensure a controlled comparison of semantic alignment and visual fidelity.}
    \Description{Twelve side-by-side text-to-image generation pairs are shown, with prompts labeled above each pair. Baseline Stable Diffusion results are on the left, and our method’s results are on the right. Our outputs exhibit stronger semantic alignment, object consistency, and perceptual fidelity than the baseline.}
    \label{fig:visual_results2}
\end{figure*}

\begin{table*}[t]
  \caption{Quantitative results on unseen text prompts across multiple datasets. Baseline denotes the original model, and Fine-tuned denotes our LoRA-based model. $\uparrow$ indicates higher values are better.}
  \label{tab:prompt_generalization}
  \centering
  \setlength{\tabcolsep}{10pt}
  \begin{tabular}{lccc|ccc|c}
    \toprule
    \textbf{Dataset} & \textbf{Metric} & \textbf{Baseline} & \textbf{Fine-tuned} & \textbf{Improvement} & \textbf{Std.} & \textbf{$t$-statistic} & \textbf{Win Rate} \\
    \midrule
    \multirow{3}{*}{DiffusionDB\cite{wang2023diffusiondb}} 
    & Pickscore $\uparrow$ & 18.8818 & 19.0554 & 0.1736 & 0.6876 & 3.5713 & 61.5\% \\
    & CLIPscore $\uparrow$ & 0.3169 & 0.3279 & 0.0110 & 0.0252 & 6.1726 & 68.5\% \\
    & IPCE $\uparrow$ & 2.6718 & 2.6792 & 0.0075 & 0.1162 & 1.7107 & 47.5\% \\
    \midrule
    \multirow{3}{*}{DrawBench\cite{Saharia2022Imagen}} 
    & Pickscore $\uparrow$ & 20.6074 & 21.0837 & 0.4763 & 0.8514 & 7.9104 & 70.5\% \\
    & CLIPscore $\uparrow$ & 0.2945 & 0.3125 & 0.0180 & 0.0334 & 7.6321 & 68.0\% \\
    & IPCE $\uparrow$ & 2.8404 & 2.8768 & 0.0364 & 0.1991 & 2.5837 & 51.5\% \\
    \midrule
    \multirow{3}{*}{PartiPrompts\cite{Yu2022Scaling}} 
    & Pickscore $\uparrow$ & 20.2817 & 21.1948 & 0.9131 & 0.9939 & 7.4635 & 83.3\% \\
    & CLIPscore $\uparrow$ & 0.2638 & 0.2982 & 0.0344 & 0.0376 & 7.4262 & 84.9\% \\
    & IPCE $\uparrow$ & 2.9349 & 3.0168 & 0.0819 & 0.2513 & 2.6468 & 63.6\% \\
    \bottomrule
  \end{tabular}
\end{table*}

To evaluate ACPO on complex text-to-image generation, we compare the baseline Stable Diffusion (SD) against our fine-tuned model on the Visual Genome dataset. Crucially, as detailed in our method, this perceptual optimization is applied exclusively during the late denoising stages ($t < 300$ out of 1,000).

\paragraph{Objective Results:}
As summarized in Table~\ref{tab:SD_main_results}, ACPO yields consistent and comprehensive gains, confirming that integrating late-stage perceptual guidance significantly elevates the generation capabilities of SD. Improvements in PickScore and CLIPScore demonstrate superior alignment with human aesthetics and textual conditions, respectively. Concurrently, increased TwoStream-IQA and IPCE scores reflect enhanced perceptual fidelity. Statistical significance is robustly confirmed via Student's $t$-tests ($t > 2$ across all metrics), while the win-rate analysis verifies that ACPO outperforms the baseline on the vast majority of samples, successfully bridging the gap between semantic accuracy and visual quality.

\paragraph{Subjective Results:}
Figure~\ref{fig:visual_results2} qualitatively validates these improvements under diverse, complex prompts. The baseline frequently struggles with fine-grained attribute binding (e.g., failing to generate the correct "hexagonal" geometry for a red stop sign) and ignores strict color constraints, occasionally collapsing into abstract textures or distorted structures. In stark contrast, ACPO explicitly resolves this attribute mismatch and structural instability. By strictly enforcing semantic alignment, generated subjects exhibit greater semantic consistency, sharper boundaries, plausible geometries, and fewer artifacts. Ultimately, this visual evidence confirms the framework yields outputs simultaneously more visually appealing and strictly faithful to complex textual prompts.

\subsection{Generalization Study}
\subsubsection{Different Text Prompts}
To evaluate whether ACPO's improvements generalize beyond the training distribution, we assess the fine-tuned model on three diverse, unseen prompt benchmarks: DiffusionDB \cite{wang2023diffusiondb}, DrawBench \cite{Saharia2022Imagen}, and PartiPrompts \cite{Yu2022Scaling}. As reported in Table~\ref{tab:prompt_generalization}, our method consistently outperforms the baseline across PickScore, CLIPScore, and IPCE. Specifically, ACPO achieves win rates exceeding 50\% for both text-image semantic alignment (CLIPScore) and human aesthetic preference (PickScore) across all three novel datasets. Supported by positive $t$-statistics that confirm statistical significance, these results robustly demonstrate that our framework generalizes its enhanced perceptual quality and strict semantic binding to completely out-of-distribution text prompts.

\subsubsection{Different Training Data Sizes}

\begin{table}[t]
  \caption{Generalization Performance on Different Training Dataset Sizes}
  \label{tab:dataset_size_ablation}
  \centering
  \setlength{\tabcolsep}{8pt} 
  \begin{tabular}{lcccc}
    \toprule
    \textbf{Dataset Size} & \textbf{Method} & \textbf{IPCE $\uparrow$} & \textbf{FID $\downarrow$} & \textbf{IS $\uparrow$} \\
    \midrule
    -            & baseline & 2.872 & - & - \\
    5k           & ours     & 2.873 & 22.36 & 12.30 \\
    10k          & ours     & 2.954 & 16.71 & 15.22 \\
    60k          & ours     & 2.969 & 16.10 & 16.77 \\
    \bottomrule
  \end{tabular}
\end{table}
To assess the data efficiency of the ACPO framework, we evaluate models fine-tuned on subsets of 5k, 10k, and 60k samples (Table~\ref{tab:dataset_size_ablation}). Notably, ACPO consistently enhances perceptual quality even in extreme low-data regimes: fine-tuning with merely 5k samples increases the baseline IPCE score from 2.872 to 2.956. Furthermore, the framework scales robustly as data increases. Expanding the training set to 60k samples further drives down the FID score from 22.36 (at 5k) to 16.10, while boosting the Inception Score (IS) from 12.30 to 16.77. These results demonstrate that our method delivers immediate perceptual improvements with limited data, while continuing to scale effectively for higher visual fidelity and generation diversity.

\subsection{Ablation study}
\subsubsection{Effectiveness of Quality-Loss}
\begin{table}
  \caption{Effectiveness of Quality Loss on DDPM}
  \label{tab:DDPM Quality-Loss}
  \centering
  \setlength{\tabcolsep}{5pt} 
  \begin{tabular}{lccc}
    \toprule
    \textbf{Model} & \textbf{Twostream $\uparrow$} & \textbf{Maniqa $\uparrow$} & \textbf{Hyper $\uparrow$} \\
    \midrule
    DDPM baseline & 82.53 & 24.16 & 0.24 \\
    DDPM Quality Loss & 84.69 & 24.21 & 0.25 \\
    \bottomrule
  \end{tabular}
\end{table}

\begin{table}
  \caption{Effectiveness of Quality Loss on SD}
  \label{tab:SD Quality-Loss}
  \centering
  \setlength{\tabcolsep}{7pt} 
  \begin{tabular}{lccc}
    \toprule
    \textbf{Model} & \textbf{IPCE $\uparrow$} & \textbf{PickScore $\uparrow$} & \textbf{CLIPScore $\uparrow$} \\
    \midrule
    SD baseline & 2.88 & 20.58 & 0.28 \\
    SD Quality Loss & 2.92 & 21.11 & 0.30 \\
    \bottomrule
  \end{tabular}
\end{table}
To isolate the impact of our perceptual objective, we evaluate the isolated Quality Loss on DDPM and SD (Tables~\ref{tab:DDPM Quality-Loss} and \ref{tab:SD Quality-Loss}). For DDPM, introducing this loss increases the TwoStream score from 82.53 to 84.69, alongside steady gains in MANIQA and HyperIQA. Similarly, applying it to SD yields consistent improvements across IPCE, PickScore, and CLIPScore. These results confirm that explicit perceptual guidance effectively enhances generation quality across diverse diffusion architectures.

\subsubsection{Quality Loss Weighting Effect}

\begin{table}[t]
  \caption{Quality Loss Weighting Effect}
  \label{tab:weight_ablation_ddpm}
  \centering
  \setlength{\tabcolsep}{10pt} 
  \begin{tabular}{lccc}
    \toprule
    \textbf{Model} & \textbf{Twostream $\uparrow$} & \textbf{Maniqa $\uparrow$} & \textbf{Hyper $\uparrow$} \\
    \midrule
    Weight 0.1 & 65.0081 & 26.45 & 0.2464 \\
    Weight 1   & 67.3500 & 23.50 & 0.2439 \\
    Weight 10  & 75.2800 & 12.61 & 0.1941 \\
    \bottomrule
  \end{tabular}
\end{table}

In this part, Table~\ref{tab:weight_ablation_ddpm} ablates the weighting coefficient of the Quality Loss. A small weight (0.1) yields marginal perceptual gains, whereas an excessively large weight (10) degrades all metrics by over-constraining the optimization and disrupting the generative prior. Setting the weight to 1.0 achieves the optimal balance, maximizing perceptual quality without compromising diffusion stability.

\subsubsection{Time Step Effect}

\begin{table}[t]
  \caption{Ablation Study on Timestep Intervention Strategies for DDPM}
  \label{tab:ablation_timestep}
  \centering
  \setlength{\tabcolsep}{0.5pt} 
  \begin{tabular}{lccc}
    \toprule
    \textbf{Method} & \textbf{TwostreamIQA $\uparrow$} & \textbf{MANIQA $\uparrow$} & \textbf{HyperIQA $\uparrow$} \\
    \midrule
    Baseline             & 78.99 & 23.50  & 0.24 \\
    Full Fine-tuning       & 82.69 & 24.81  & 0.19 \\
    Fine-tune Last 1 Step & 80.02 & 23.35  & 0.24 \\
    Fine-tune Last 10 Steps & 80.21 & 24.21 & 0.25 \\
    \bottomrule
  \end{tabular}
\end{table}
We ablate the timestep intervention strategy by comparing full-trajectory fine-tuning against targeted late-stage optimization (last 10 steps vs. last 1 step) in Table~\ref{tab:ablation_timestep}. While full fine-tuning achieves the highest TwoStream and MANIQA (24.81) scores, applying the loss exclusively to the final 10 denoising steps yields a more stable balance, even surpassing full fine-tuning in HyperIQA. Conversely, intervening only at the final step provides negligible benefits. This validates our core ACPO design: targeting a concise window of late denoising stages efficiently solidifies perceptual details without the instability of full-trajectory optimization.

\section{Conclusion}
In this paper, we provide a novel insight into diffusion model training: generated images must not only achieve high pixel-level reconstruction fidelity, but also exhibit superior perceptual quality and text-image semantic consistency. Recognizing that directly injecting no-reference perceptual signals often leads to optimization mismatch and distributional drift during fine-tuning, we propose an effective anchor-constrained optimization framework. Within this framework, we utilize a learned no-reference image quality assessment (NR-IQA) model to provide direct perceptual guidance, driving generation toward perceptually favorable outputs. Crucially, to prevent the model from forgetting its original generative capabilities, we introduce an anchor-based regularization mechanism that enforces consistency with the frozen base diffusion model in noise prediction. This proposed strategy effectively balances the pursuit of superior visual perception and text-image semantic consistency with the retention of generative fidelity. Extensive experiments demonstrate that our method consistently improves the perceptual quality of generated images across multiple datasets, while preserving generation diversity and ensuring training stability, thereby highlighting the effectiveness of anchor-constrained perceptual adaptation for diffusion models.


\bibliographystyle{ACM-Reference-Format}
\bibliography{references}

\clearpage
\appendix
\section*{Supplementary Material}
\addcontentsline{toc}{section}{Supplementary Material}
This supplementary material complements the main paper by providing an expanded set of visual evaluations and supplementary experimental analyses. Section 1 presents extended qualitative comparisons for our baseline experiments, offering further insights into the perceptual strengths of our approach. Section 2 showcases a broader range of visual examples illustrating the method's generalization capabilities across different datasets and diverse text prompts. Finally, Section 3 presents extended experimental analyses, including more comprehensive ablation studies conducted on the Stable Diffusion (SD) architecture to evaluate the specific impacts of timestep intervention, quality weighting, and anchor loss.

\section{Extended Qualitative Results}
This section provides additional qualitative evidence to support the effectiveness of the Anchor-Constrained Perceptual Optimization (ACPO) framework. Following the logic established in the main paper, we showcase results across two dimensions: general perceptual quality enhancement on unconditional models and semantic alignment refinement on text-to-image models.

\subsection{Additional Visual Evidence for Perceptual Quality}
\begin{figure*}[htbp]
    \centering
    \includegraphics[width=\textwidth]{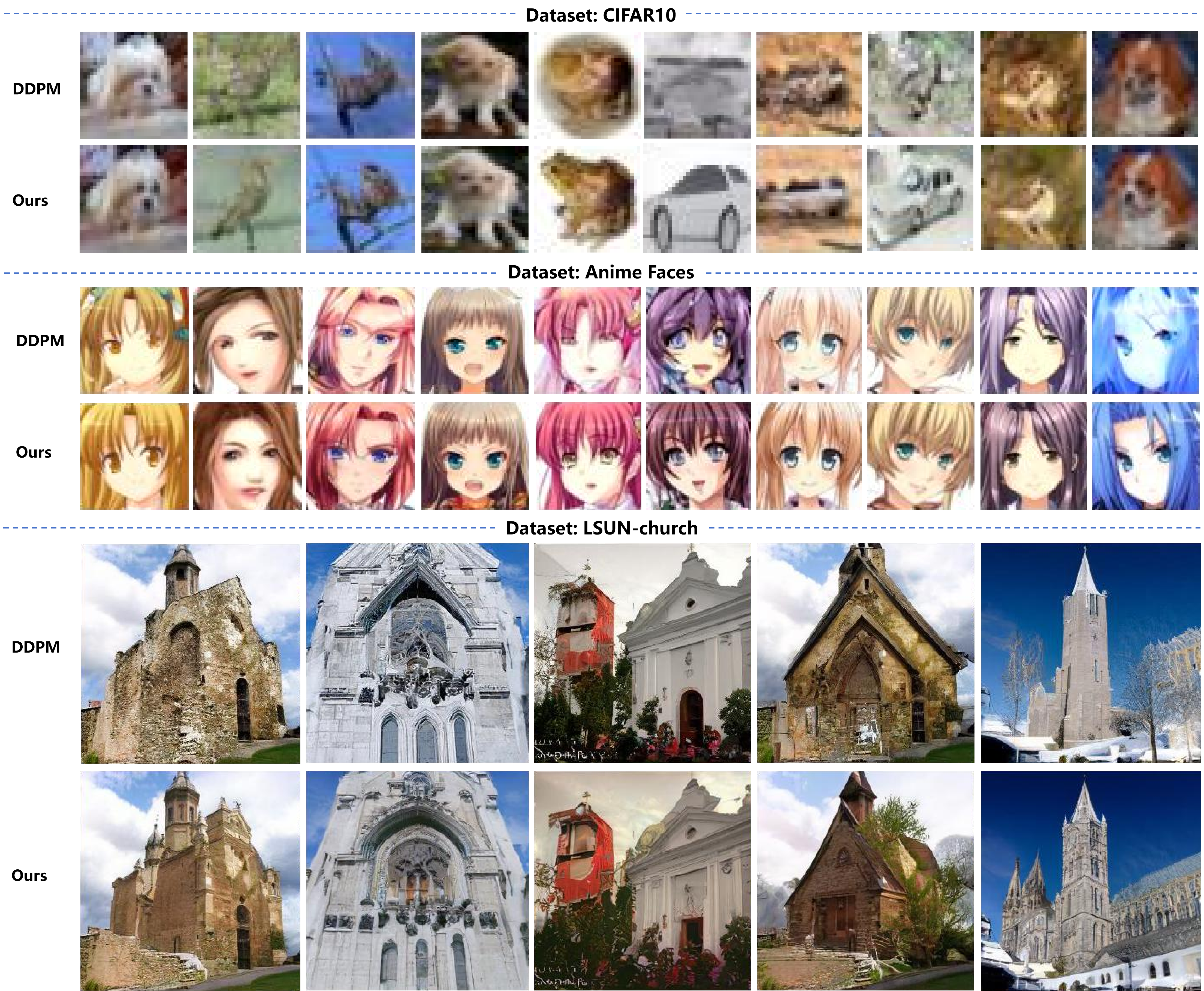}
    \caption{Extended qualitative comparison between the baseline DDPM and our ACPO model across multiple datasets: CIFAR-10 ($32 \times 32$), Anime Faces ($64 \times 64$), and LSUN-church ($256 \times 256$). For each dataset, images in the top row are generated by the baseline, while the bottom row shows the results after our iterative optimization. Each column is synthesized from the same initial Gaussian noise under identical sampling parameters to ensure a controlled and fair evaluation of perceptual refinements.}
    \Description{The figure presents three groups of image comparisons across different datasets. In each group, the upper row shows baseline results and the lower row shows our results. For CIFAR-10, our method produces objects with more defined boundaries. For Anime Faces, the results exhibit more vivid colors and cleaner facial features. For the LSUN-church dataset, our framework synthesizes architectural structures with enhanced textural detail and sharper contrast compared to the baseline's plausible but softer outputs.}
    \label{fig:DDPM_extend}
\end{figure*}
Figure~\ref{fig:DDPM_extend} presents a comparative visualization between the baseline DDPM and our ACPO-enhanced results across three diverse datasets: CIFAR-10, Anime Faces, and LSUN-church. As illustrated, the baseline model effectively captures the underlying data distribution and successfully synthesizes plausible image structures. Building upon this strong generative foundation, our ACPO framework provides further refinements in perceptual clarity and fine-grained textures.

On the CIFAR-10 dataset, our method yields sharper object boundaries and more consistent color transitions. For Anime Faces and LSUN-church, the optimization further enhances the structural coherence of facial features and architectural geometries, respectively. These results demonstrate that by integrating NR-IQA signals during the late-stage denoising process, our iterative optimization effectively complements the base model's capabilities. This "polishing" effect allows the final outputs to achieve a higher degree of visual fidelity and detail, aligning them more closely with human perceptual preferences without disrupting the established generative behavior of the baseline.

\subsection{Additional Visual Evidence for Semantic Alignment}
\begin{figure*}[htbp]
    \centering
    \includegraphics[width=\textwidth]{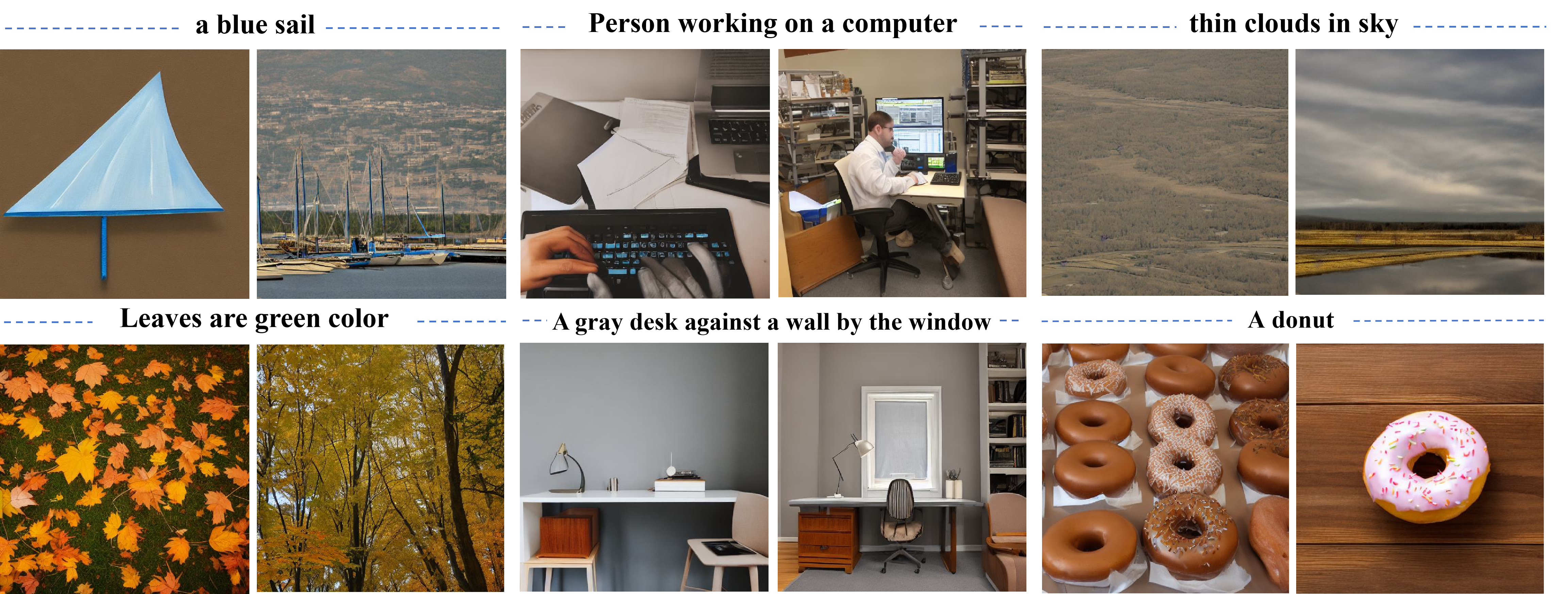}
    \caption{Visual demonstration of improved semantic alignment on Stable Diffusion. Specific text prompts are indicated by the blue dashed labels above each pair. Within each pair, the baseline Stable Diffusion output is on the left, and our ACPO result is on the right. By maintaining the same random seed for each comparison, we illustrate how our framework further guides the model to produce images that more accurately reflect the descriptive attributes of the prompts.}
    \Description{Six sets of side-by-side comparisons show text-to-image generation results. For the prompt 'Leaves are green color', our model successfully renders green foliage, demonstrating precise color attribute binding. In the 'A donut' example, our method produces a single, high-fidelity pink donut with realistic textures, whereas 'A gray desk against a wall by the window' shows a cleaner, more organized interior layout. Across all examples, such as 'thin clouds in sky' and 'a blue sail', our method exhibits enhanced adherence to the semantic details provided in the text, resulting in more contextually grounded visual outputs.}
    \label{fig:SD_extend}
\end{figure*}
Beyond basic visual quality, ACPO demonstrates a unique ability to reinforce text-image semantic alignment when the guiding metric is semantic-aware. Figure~\ref{fig:SD_extend} presents qualitative comparisons using the Stable Diffusion (SD) architecture on a variety of descriptive prompts. While the baseline establishes a solid generative foundation, the integration of semantic-aware perceptual guidance allows our framework to achieve a higher degree of precision in attribute binding and contextual grounding. For instance, in the prompt "Leaves are green color", our method more accurately captures the specified color attribute, ensuring the visual output remains faithful to the text. Similarly, for prompts such as "A gray desk against a wall by the window" and "A donut", our approach results in clearer semantic interpretation and more distinct object representation. Even for subtle descriptions like "thin clouds in sky", the ACPO framework further refines the texture to match the expectation. These comparisons highlight that ACPO effectively leverages semantic signals to steer the generation process, ensuring the final outputs are more closely aligned with the conceptual intent of the user-provided text.

\section{Visual Results for Generalization}
This section provides an expanded set of visual results on several unseen prompt benchmarks—DrawBench, DiffusionDB, and PartiPrompts—to further validate the generalization capability of the ACPO framework. These qualitative comparisons serve as a visual counterpart to the quantitative findings reported in the main text, demonstrating that our method effectively transfers its perceptual and semantic benefits to completely out-of-distribution prompts.
\begin{figure*}[t]
    \centering
    \includegraphics[width=\textwidth]{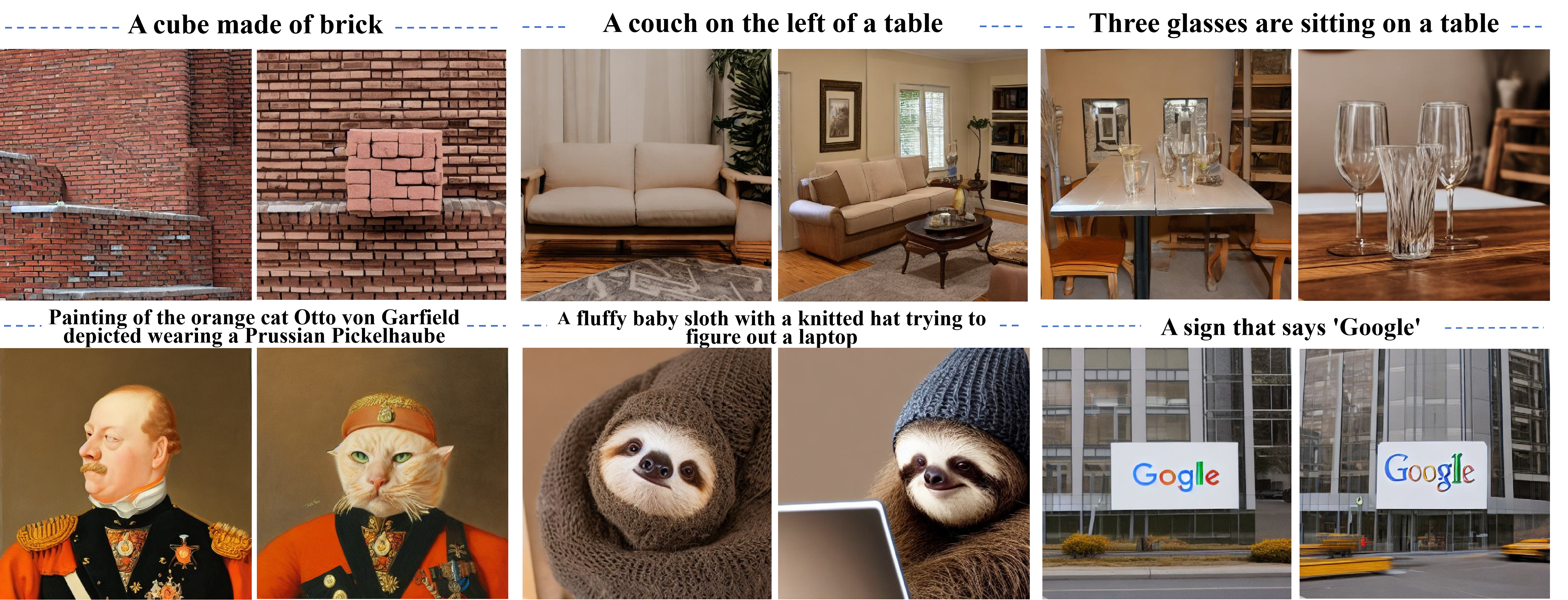}
    \caption{Qualitative results on the DrawBench benchmark. For each comparison, the baseline Stable Diffusion result is shown on the left and our ACPO result is on the right. By utilizing identical sampling settings, we demonstrate the framework's ability to achieve better alignment with challenging structural and relational descriptions.}
    \Description{Six pairs of images from DrawBench are displayed. In examples such as 'A couch on the left of a table' and 'Three glasses are sitting on a table', our ACPO framework produces results with more accurate spatial positioning and count adherence. For the 'Otto von Garfield' prompt, the refined model successfully captures the specific historical attributes while maintaining a high degree of character consistency compared to the baseline.}
    \label{fig:DrawBench}
\end{figure*}
\begin{figure*}[t]
    \centering
    \includegraphics[width=\textwidth]{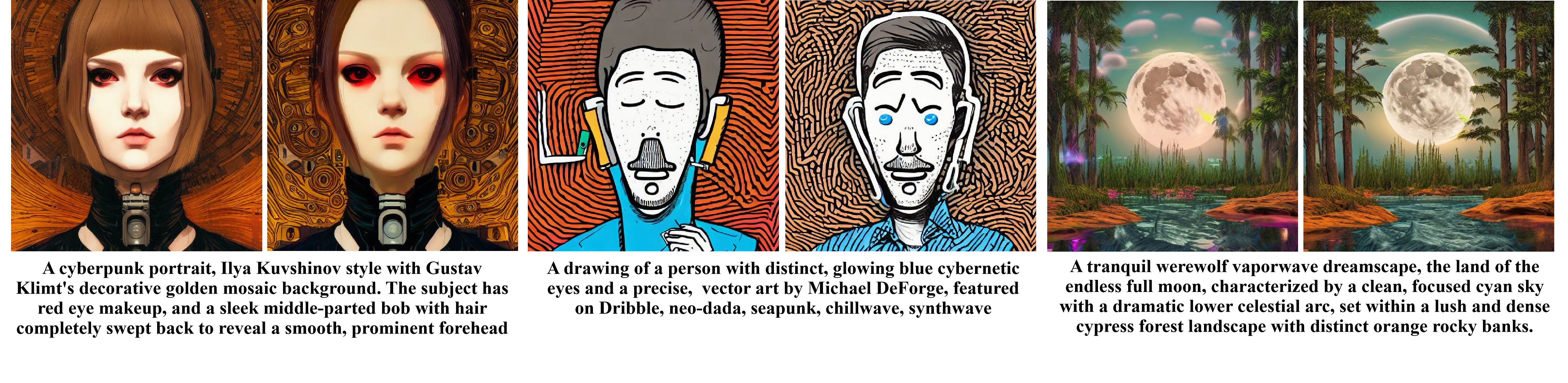}
    \caption{Generalization results on the PartiPrompts benchmark across various common object categories. Each pair consists of a baseline result (left) and our ACPO-enhanced result (right), generated from the same initial noise. These comparisons demonstrate the consistent effectiveness of our framework in refining structural integrity and textural clarity across a diverse range of unseen semantic domains.}
    \Description{Three pairs of high-detail images are shown. For the cyberpunk portrait, our method successfully reveals a 'smooth, prominent forehead' and clean hair styling as requested by the prompt. In the vector art and vaporwave dreamscape examples, ACPO demonstrates a more precise rendering of distinctive artistic textures and celestial elements, resulting in a more cohesive visual interpretation of the complex prompts.}
    \label{fig:DiffusionDB}
\end{figure*}
\begin{figure*}[t]
    \centering
    \includegraphics[width=\textwidth]{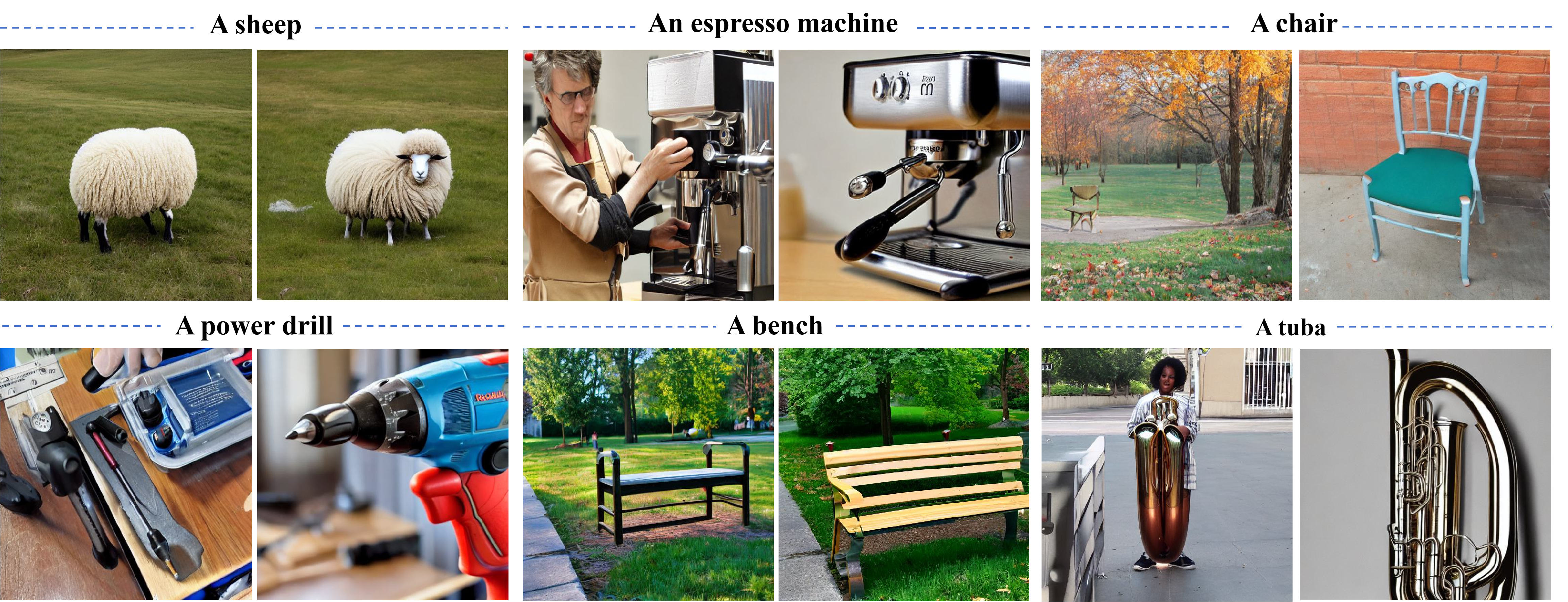}
    \caption{Generalization results on the PartiPrompts benchmark across various common object categories. Each pair consists of a baseline result (left) and our ACPO-enhanced result (right), generated from the same initial noise. These comparisons showcase the robustness of our framework in refining the structural integrity and textural clarity of diverse subjects in out-of-distribution scenarios.}
    \Description{Six pairs of images show common objects including a sheep, an espresso machine, a chair, a power drill, a bench, and a tuba. In each case, the ACPO-refined images exhibit cleaner details and more defined structures. For example, the complex mechanical parts of the 'espresso machine' and 'power drill' are rendered with greater clarity, and the 'tuba' shows enhanced metallic reflections and geometric accuracy compared to the baseline.}
    \label{fig:PartiPrompts}
\end{figure*}
Figure~\ref{fig:DrawBench} displays results from the DrawBench benchmark. The comparisons highlight the model's improved capacity to synthesize scenes involving specific spatial arrangements and object quantities, such as the precise placement of furniture or the correct count of items on a table. In Figure~\ref{fig:DiffusionDB}, we provide additional examples from DiffusionDB, where the ACPO-refined model demonstrates an enhanced ability to ground intricate artistic styles and specific visual attributes. Whether handling decorative mosaic patterns or detailed character features, the optimized model ensures a closer match to the descriptive prompts while maintaining high textural fidelity. Finally, Figure~\ref{fig:PartiPrompts} illustrates results on PartiPrompts, showcasing the model's robustness across a wide variety of common objects and general scenarios. Across these diverse benchmarks, our framework consistently produces cleaner, more focused representations with enhanced structural integrity. Collectively, these results confirm that ACPO provides a robust and generalizable optimization signal that maintains significant performance gains across a broad spectrum of semantic domains.

\section{Comprehensive Ablation Analysis on Stable Diffusion}
To further validate the robustness and sensitivity of the Anchor-Constrained Perceptual Optimization (ACPO) framework, we conduct an extensive ablation study on the Stable Diffusion (SD) architecture. All experiments in this section are conducted on a controlled subset of 5,000 images sampled from the Visual Genome dataset. The models are fine-tuned for a single epoch to observe the immediate impact of perceptual guidance and anchor regularization.

\subsection{Effectiveness of Anchor-Based Regularization}
\begin{table}[t]
  \caption{Ablation Study on Loss Components}
  \label{tab:loss_sd}
  \centering
  \setlength{\tabcolsep}{1.5pt}
  \begin{tabular}{lcccc}
    \toprule
    \textbf{Method} & \textbf{IPCE $\uparrow$} & \textbf{CLIPScore $\uparrow$} & \textbf{PickScore $\uparrow$} & \textbf{Twostream $\uparrow$} \\
    \midrule
    Baseline      & 2.8130 & 0.2980 & 18.3066 & 75.3494 \\
    w/ Quality   & 2.9111 & 0.2859 & 18.1289 & 64.0657 \\
    w/ Anchor(ours)   & 2.9051 & 0.3032 & 18.3877 & 78.8847 \\
    \bottomrule
  \end{tabular}
\end{table}
We first investigate the necessity of the anchor-based regularization mechanism. In this setup, we fix the intervention window to the last 100 time steps ($t < 100$) and set the quality loss weight to 0.1. As reported in Table~\ref{tab:loss_sd}, optimizing the perceptual signal in isolation ("Only Quality Loss") leads to a substantial improvement in the IPCE score (2.9111) but results in a catastrophic drop in the TwoStream-IQA score (64.06) and a decline in semantic metrics like CLIPScore and PickScore. This indicates that without the anchor constraint, the model suffers from severe distributional drift. Conversely, our ACPO formulation successfully harnesses the perceptual gains while maintaining—and even enhancing—generative stability and semantic alignment, with the TwoStream score rising to 78.88.

\subsection{Influence of Perceptual Quality Weighting}
\begin{table}[t]
  \caption{Ablation Study on Quality Loss Weight}
  \label{tab:weight_sd}
  \centering
  \setlength{\tabcolsep}{1.5pt}
  \begin{tabular}{lcccc}
    \toprule
    \textbf{Quality Weight} & \textbf{IPCE $\uparrow$} & \textbf{CLIPScore $\uparrow$} & \textbf{PickScore $\uparrow$} & \textbf{Twostream $\uparrow$} \\
    \midrule
    Baseline & 2.8130 & 0.2980 & 18.3066 & 75.3494 \\
    0.1      & 2.9051 & 0.3032 & 18.3877 & 78.8847 \\
    1        & 2.9445 & 0.2907 & 18.5117 & 72.7214 \\
    100      & 3.1156 & 0.2966 & 18.4013 & 63.1967 \\
    \bottomrule
  \end{tabular}
\end{table}
We evaluate the sensitivity of the optimization to the quality loss weighting coefficient, with the timestep window fixed at $t < 100$. Table~\ref{tab:weight_sd} demonstrates a clear trade-off between perceptual refinement and generative prior preservation. A moderate weight (0.1 to 1.0) achieves the most balanced performance, yielding consistent improvements across all metrics. While a high weight (100) significantly boosts the IPCE score to 3.1156, it simultaneously compromises the TwoStream stability score (63.19). This confirms that a carefully calibrated weighting is essential to ensure that the perceptual guidance complements rather than disrupts the pre-trained diffusion backbone.

\subsection{Evaluation of Timestep Intervention Ranges}
\begin{table}[t]
  \caption{Ablation Study on Intervention Time Steps}
  \label{tab:timestep_sd}
  \centering
  \setlength{\tabcolsep}{2pt}
  \begin{tabular}{lcccc}
    \toprule
    \textbf{TimeStep ($t <$)} & \textbf{IPCE $\uparrow$} & \textbf{CLIPScore $\uparrow$} & \textbf{PickScore $\uparrow$} & \textbf{Twostream $\uparrow$} \\
    \midrule
    Baseline & 2.8130 & 0.2980 & 18.3066 & 75.3494 \\
    100      & 2.9051 & 0.3032 & 18.3877 & 78.8847 \\
    300      & 2.9578 & 0.3083 & 18.5117 & 78.7808 \\
    500      & 2.9660 & 0.2922 & 18.4794 & 76.0279 \\
    1000     & 2.9533 & 0.2881 & 18.4687 & 56.2540 \\
    \bottomrule
  \end{tabular}
\end{table}
Finally, we ablate the optimal window for iterative optimization by varying the maximum timestep $t$, while keeping the quality weight fixed at 0.1. As shown in Table~\ref{tab:timestep_sd}, targeting the late denoising stages ($t < 100$ or $t < 300$) provides the most robust gains in both semantic alignment and visual quality. Notably, extending the optimization to the entire trajectory ($t < 1000$) results in a significant collapse of the TwoStream score (56.25) and a decline in semantic consistency (CLIPScore 0.2881). These findings empirically support our design choice: by concentrating the perceptual optimization on the final denoising stages where textures solidify, ACPO effectively refines visual details without inducing the instability inherent in full-trajectory fine-tuning.

\end{document}